\documentclass[3p,twocolumn,times]{elsarticle}

\usepackage{amssymb}
\usepackage{amsmath, amsfonts, graphicx, booktabs, tabularx, multirow, siunitx, xcolor, soul, bm, enumitem}
\usepackage[colorlinks = true,
            linkcolor = blue,
            urlcolor  = blue,
            citecolor = blue,
            anchorcolor = blue]{hyperref}
\usepackage{algpseudocode}
\usepackage{algorithm}

\algnewcommand\algorithmicforeach{\textbf{for each}}
\algdef{S}[FOR]{ForEach}[1]{\algorithmicforeach\ #1\ \algorithmicdo}

\algnewcommand{\LineComment}[1]{\State \(\triangleright\) #1}

\graphicspath{{./figures/}}

\journal{Pattern Recognition Letters}

\begin{document}

\begin{frontmatter}



\title{Basis Scaling and Double Pruning for Efficient Inference in Network-Based Transfer Learning\tnoteref{t1}}
\tnotetext[t1]{This paper was accepted by Pattern Recognition Letters (\url{https://doi.org/10.1016/j.patrec.2023.11.026})}


%

\author[label1]{Ken C. L. Wong\corref{cor1}}
\ead{clwong@us.ibm.com}
\author[label1]{Satyananda Kashyap}
\ead{satyananda.kashyap@ibm.com}
\author[label2]{Mehdi Moradi}
\ead{moradm4@mcmaster.ca}

\affiliation[label1]{
organization={IBM Research -- Almaden Research Center},
city={San Jose},
state={CA},
country={USA}}

\affiliation[label2]{
organization={McMaster University},
city={Hamilton},
state={ON},
country={Canada}}

\cortext[cor1]{Corresponding author.}

\begin{abstract}
Network-based transfer learning allows the reuse of deep learning features with limited data, but the resulting models can be unnecessarily large. Although network pruning can improve inference efficiency, existing algorithms usually require fine-tuning that may not be suitable for small datasets. In this paper, using the singular value decomposition, we decompose a convolutional layer into two layers: a convolutional layer with the orthonormal basis vectors as the filters, and a ``BasisScalingConv'' layer which is responsible for rescaling the features and transforming them back to the original space. As the filters in each decomposed layer are linearly independent, when using the proposed basis scaling factors with the Taylor approximation of importance, pruning can be more effective and fine-tuning individual weights is unnecessary. Furthermore, as the numbers of input and output channels of the original convolutional layer remain unchanged after basis pruning, it is applicable to virtually all architectures and can be combined with existing pruning algorithms for double pruning to further increase the pruning capability. When transferring knowledge from ImageNet pre-trained models to different target domains, with less than 1\% reduction in classification accuracies, we can achieve pruning ratios up to 74.6\% for CIFAR-10 and 98.9\% for MNIST in model parameters.
\end{abstract}


\end{frontmatter}


\section{Introduction}\label{sec:introduction}


Deep convolutional neural networks have dominated the area of applied computer vision. The network architectures used for image analysis have grown in terms of performance along with their numbers of layers and parameters over the years. Nevertheless, as the ubiquitous use of these networks is now extended to resource-limited areas such as edge computing, optimization of the architectures to minimize computational requirements has become essential. Furthermore, the reduction in floating point operations (FLOPs) at inference time directly impacts the power consumption of the large-scale consumer facing applications of artificial intelligence. Thus, the advocates of green AI recommend the use of network size and the number of FLOPs as important performance evaluation metrics for neural networks, along with accuracy \cite{Journal:Schwartz:CACM2020}.


To achieve the goal of computational efficiency, a significant amount of progress has been made in the area of pruning. Pruning is the process of finding the components of a network that can be removed without a large loss of performance. Since the early days of convolutional neural networks, it has been shown that removing unimportant components delivers benefits in generalization and computational efficiency \cite{Conference:Lecun:NIPS1990}.


Similar to \cite{Conference:Molchanov:ICLR2017:Pruning}, we focus on pruning in the context of transfer learning. The goal of transfer learning is to transfer knowledge from a source domain to a target domain, where the data distributions in the two domains are different \cite{Journal:Pan:TKDE2010:survey,Conference:Tan:AAAI2017:distant,Conference:Tan:ICANN2018:Survey,Journal:Huang:PR2020:transfer}. Transfer learning is necessary for the areas in which large-scale and well-annotated datasets are scarce due to the cost of data acquisition and annotation. While there are different approaches of transfer learning \cite{Conference:Tan:ICANN2018:Survey}, such as instances-based and mapping-based approaches that require the availability of the source data, here we study the network-based transfer learning which allows the transfer of knowledge learned from millions of samples (e.g., ImageNet \cite{Conference:Deng:CVPR2009}) without the difficulties of handling the enormous source data. Furthermore, instead of domain adaptation which has been the focus of most work in transfer learning, we focus on the neglected problem of efficient inference in the target domain through pruning. As the source dataset usually contains features that are redundant in the target dataset, using pre-trained networks without pruning can result in unnecessarily large models. In contrast, transfer learning with pruning can achieve substantial reductions in parameters and FLOPs\footnote{In this paper, the acronym FLOPs represents the floating point operations required to infer a single instance. FLOPs are proportional to the inference time while allowing machine-independent comparison.}.


Existing network pruning algorithms are mainly achieved by removing structural contents such as filters or layers to produce weight matrices that are much smaller in size for substantial practical improvement in efficiency. While some frameworks perform pruning without considering image data \cite{Conference:Jaderberg:BMVC2014,Conference:Li:ICLR2017:Pruning,Conference:Wang:CVPR2021:convolutional,Journal:Zhang:AI2022:group}, most works use image data for better pruning ratios and accuracy \cite{Conference:Wen:NIPS2016:Learning,Conference:He:ICCV2017,Conference:Liu:ICCV2017,Conference:Luo:ICCV2017,Conference:Molchanov:ICLR2017:Pruning,Conference:Ye:ICLR2018:Rethinking,Conference:Molchanov:CVPR2019,Conference:Lin:CVPR2020}. Although these frameworks provide promising results, as discussed in \cite{Conference:Kornblith:MLR2019,Conference:Raghu:NIPS2017}, since the filters are linearly dependent in a layer, pruning in the original filter space can be less effective. Fine-tuning of the entire network after pruning is also required in most frameworks, which may not be desirable for transfer learning with limited data.

In view of these issues, here we propose a framework that fine-tunes and prunes the orthonormal bases obtained by applying the singular value decomposition (SVD) on the convolutional weight matrices. The SVD decomposes a linear transformation matrix into two sets of orthonormal bases and the corresponding singular values that indicate the contributions of the basis vectors \cite{Journal:Stewart:SIAM1993:early}. Our contributions include:
\begin{itemize}
\item
Using the linear independence of basis vectors, we propose a basis pruning algorithm that prunes convolutional layers regardless of the network architecture. As the basis vectors are non-trainable in our framework to facilitate transfer learning, we introduce the basis scaling factors which are responsible for importance estimation and fine-tuning of basis vectors. These basis scaling factors are trainable by backpropagation and only contribute a small amount of trainable parameters.
\item
After basis pruning, as the numbers of input and output channels of the original convolutional layers remain unchanged, other pruning algorithms can be further applied for double pruning. By combining the advantages of basis pruning and other pruning algorithms, we can achieve larger pruning ratios that cannot be achieved by either alone. This provides a new approach that can amplify existing pruning mechanism.
\end{itemize}
We tested our framework by transferring the features of ImageNet pre-trained models to classify the CIFAR-10, MNIST, and Fashion-MNIST datasets.

\section{Related Work}

\textbf{Channel/Filter Pruning.} Network pruning can be achieved by pruning channels/filters based on the structured sparsity. In \cite{Conference:Li:ICLR2017:Pruning}, the L1 norm of each filter was computed as its relative importance within a convolutional layer, and filters with smaller L1 norms were pruned. In \cite{Conference:He:ICCV2017,Conference:Luo:ICCV2017}, the channels were pruned while minimizing the reconstruction error between the original and modified feature maps in each layer. In \cite{Conference:Liu:ICCV2017,Conference:Ye:ICLR2018:Rethinking}, the scaling factors in the batch normalization (BN) layers were used for channel pruning. In \cite{Conference:Molchanov:CVPR2019}, the importance of a parameter was quantified by the error induced by removing it. Using Taylor expansions to approximate such errors with the gradients of the minibatches, less important filters were pruned. In \cite{Conference:Wang:CVPR2021:convolutional}, by building a graph with each vertex as a filter and the edges defined with the distances between filter weights, the structural redundancy can be quantified for pruning.

\textbf{Pruning and Matrix Factorization.} Matrix factorization techniques such as SVD and principal component analysis (PCA) have been applied to deep learning to reveal properties that cannot be observed in the original space. In \cite{Workshop:Yang:CVPRWorkshops2020}, without pre-training, the weight matrices were factorized by SVD. SVD training was then performed on the decomposed weights with orthogonality regularization and sparsity-inducing regularizers, and the resulting network was pruned by the singular values and fine-tuned. In \cite{Journal:Garg:Access2020}, PCA was applied on the feature maps to analyze the network structure for the optimal layer-wise widths and number of layers, and the resulting structure was then trained from scratch. In \cite{Conference:Lin:CVPR2020}, SVD was used to obtain the ranks of feature maps and filters with lower ranks were pruned. In \cite{Journal:Zhang:AI2022:group}, kernel-PCA was used to cluster the filters in a layer into groups, and pruning was applied in each group to replace regular convolution by grouped convolution.

\section{Methodology}

In our framework, the SVD is used to decompose a convolutional layer into two consecutive layers with the weights as orthogonal bases. By introducing the basis scaling factors, basis pruning can be achieved by computing the Taylor approximation of importance of these factors for more effective pruning. Basis pruning can be combined with existing pruning algorithms for double pruning to further increase the pruning capability.

\subsection{Convolutional Weights Representation with Orthonormal Bases}
\label{sec:weights_representation}

Let $\mathbb{W} \in \mathbb{R}^{k_h \times k_w \times c_i \times c_o}$ be a 4-D weight matrix of a convolutional layer, with $k_h$ and $k_w$ the kernel height and width, and $c_i$ and $c_o$ the numbers of input and output channels. For efficient transfer learning, we assume that the convolutional weights are pre-trained and non-trainable. $\mathbb{W}$ can be reshaped into a 2-D matrix $\mathbf{W} \in \mathbb{R}^{k \times c_o}$ for further processing ($k = k_h \times k_w \times c_i$), which can be factorized by the compact SVD as:
\begin{align}
\label{eq:svd}
\mathbf{W} = \mathbf{U} \mathbf{\Sigma} \mathbf{V}^\mathrm{T}
\end{align}
where $\mathbf{U} \in \mathbb{R}^{k \times r}$ and $\mathbf{V} \in \mathbb{R}^{c_o \times r}$ contain the columns of left-singular vectors and right-singular vectors, respectively. $\mathbf{\Sigma} \in \mathbb{R}^{r \times r}$ is a diagonal matrix of singular values in descending order. $r = \min\{k, c_o\}$ is the maximum rank of $\mathbf{W}$. As the columns of $\mathbf{U}$ yield an orthonormal basis, like those of $\mathbf{V}$, we have $\mathbf{U}^\mathrm{T} \mathbf{U} = \mathbf{V}^\mathrm{T} \mathbf{V} = \mathbf{I}$, with $\mathbf{I} \in \mathbb{R}^{r \times r}$ an identity matrix. With SVD, we can rescale and prune the orthonormal bases $\mathbf{U}$ and $\mathbf{V}$.

\subsection{Convolutional Layer Decomposition}
\label{sec:layer_decomposition}

\begin{figure}[t]
    \centering
    \includegraphics[width=0.9\linewidth]{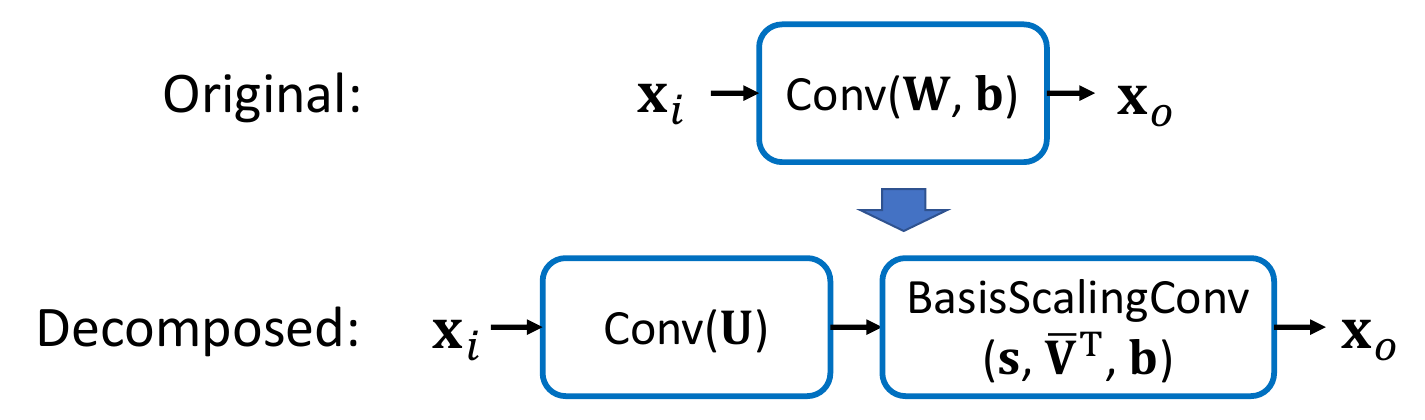}
    \caption{Decomposition of a convolutional layer using (\ref{eq:factorized_scaled_conv}). Only the vector of basis scaling factors $\mathbf{s}$ is trainable during transfer learning.}
    \label{fig:conv_decomposition}
\end{figure}

Using SVD, the convolutional weights can be represented by the orthonormal bases $\mathbf{U}$ and $\mathbf{V}$. Although the contributions of the basis vectors are proportional to the corresponding singular values, most singular values are of similar magnitudes and choosing which to be removed is nontrivial especially without considering the image data. As we also want to preserve the original weights while pruning, here we introduce the basis-scaling convolutional (\emph{BasisScalingConv}) layer to measure the importance of the basis vectors.

\begin{figure*}[t]
    \scriptsize
    \centering
    \begin{minipage}[t]{0.19\linewidth}
        \centering{\phantom{aaaa}VGG-16 layer \#2}
        \centering
        \includegraphics[width=1\linewidth]{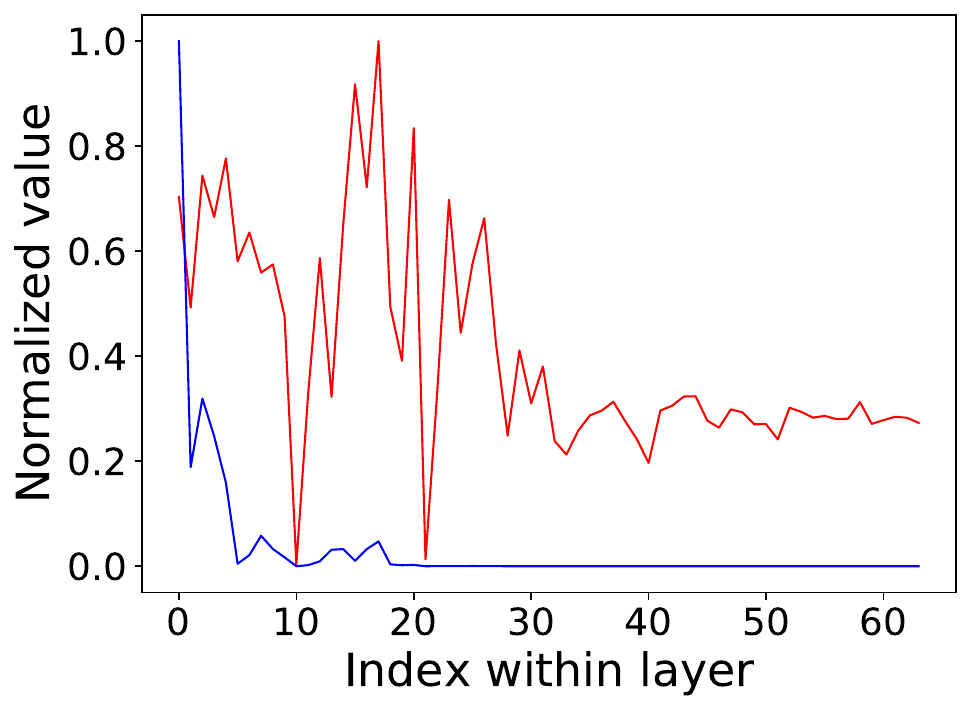}
    \end{minipage}
    \begin{minipage}[t]{0.19\linewidth}
        \centering{\phantom{aaaa}VGG-16 layer \#4}
        \centering
        \includegraphics[width=1\linewidth]{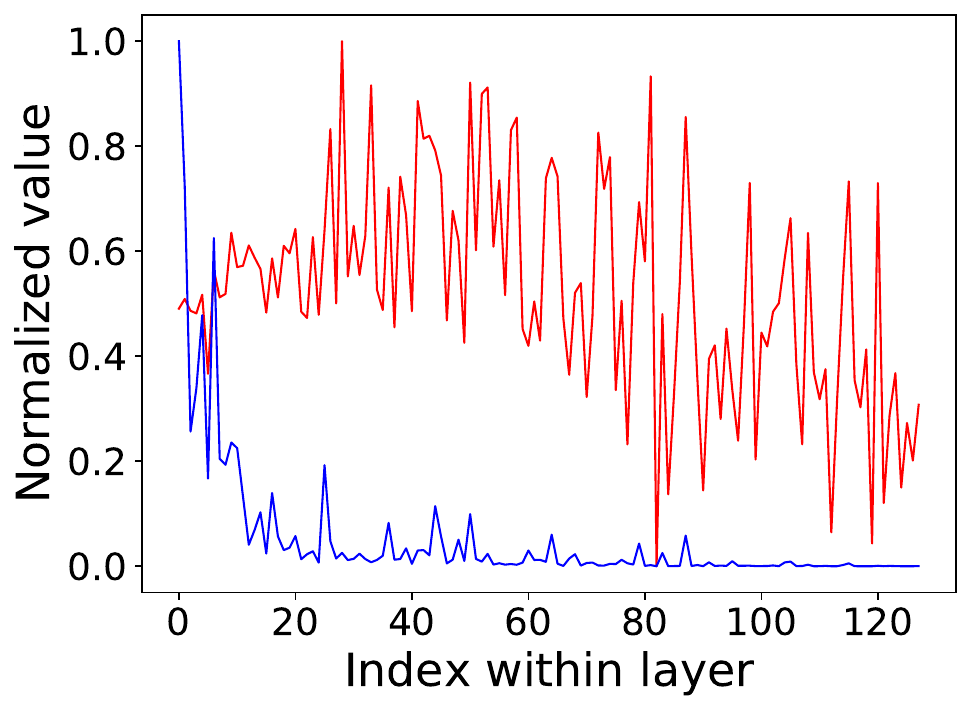}
    \end{minipage}
    \begin{minipage}[t]{0.19\linewidth}
        \centering{\phantom{aaaa}VGG-16 layer \#7}
        \centering
        \includegraphics[width=1\linewidth]{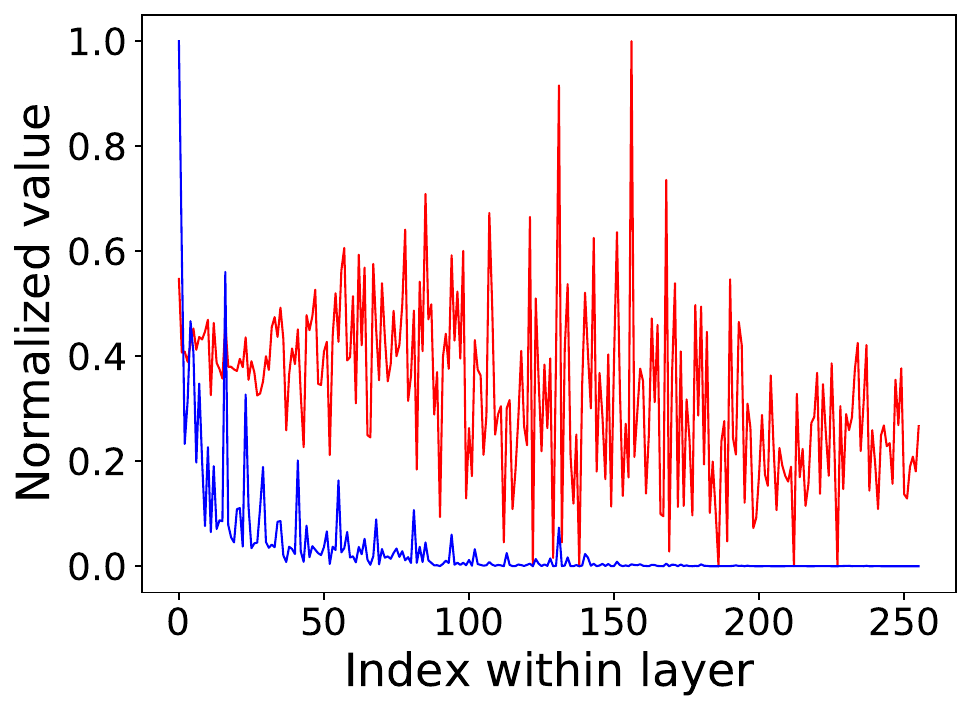}
    \end{minipage}
    \begin{minipage}[t]{0.19\linewidth}
        \centering{\phantom{aaaa}VGG-16 layer \#10}
        \centering
        \includegraphics[width=1\linewidth]{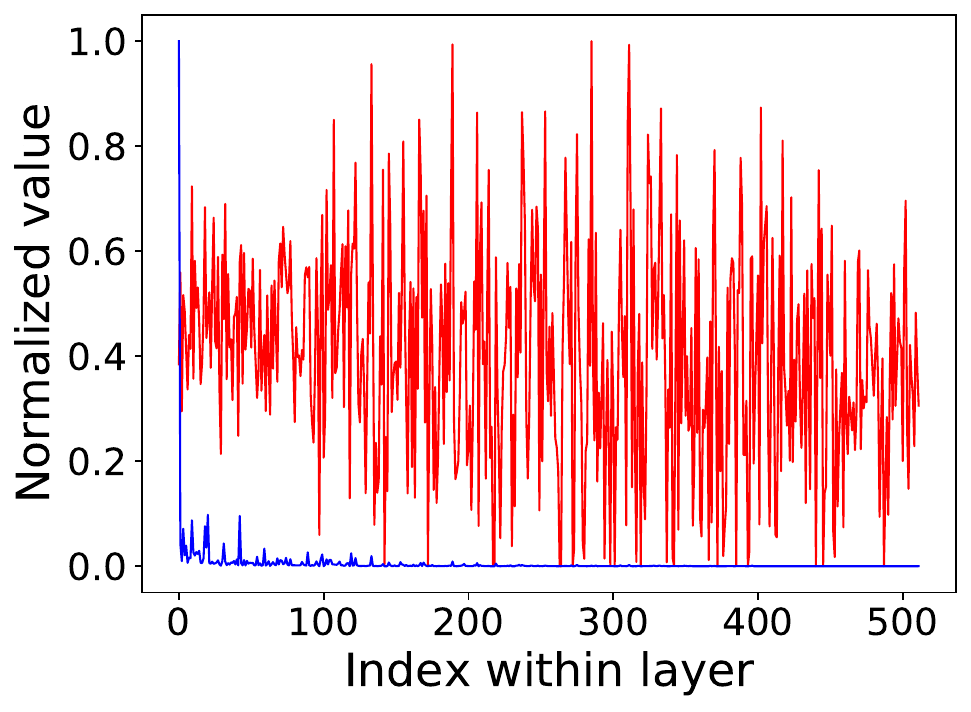}
    \end{minipage}
    \begin{minipage}[t]{0.19\linewidth}
        \centering{\phantom{aaaa}VGG-16 layer \#13}
        \centering
        \includegraphics[width=1\linewidth]{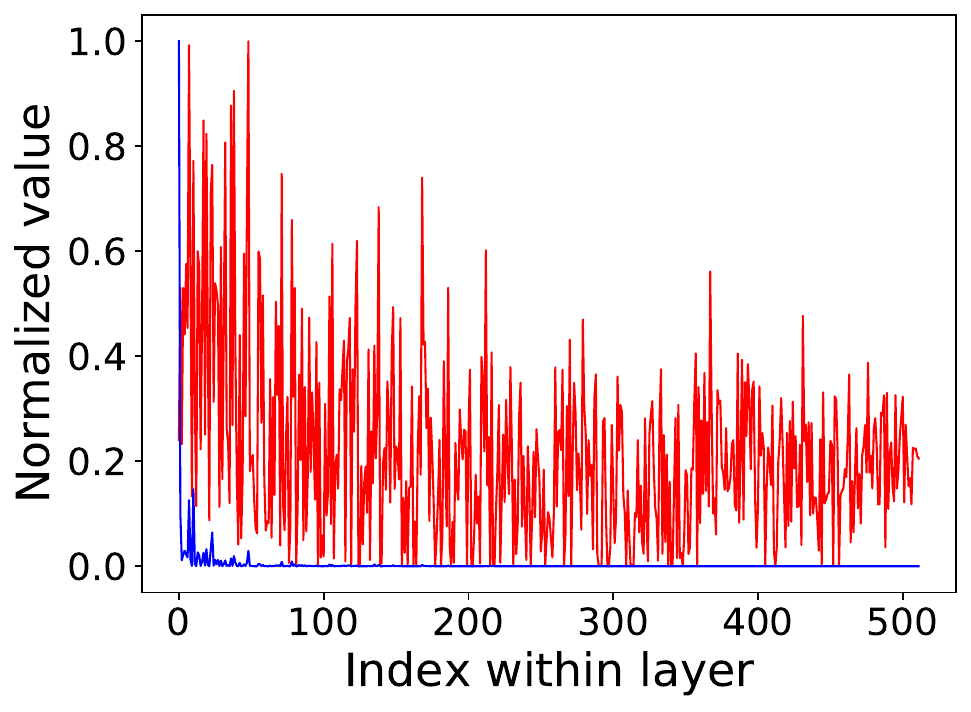}
    \end{minipage}
    \\
    \smallskip
    \begin{minipage}[t]{0.19\linewidth}
        \centering{\phantom{aaaa}DenseNet-121 layer \#1}
        \centering
        \includegraphics[width=1\linewidth]{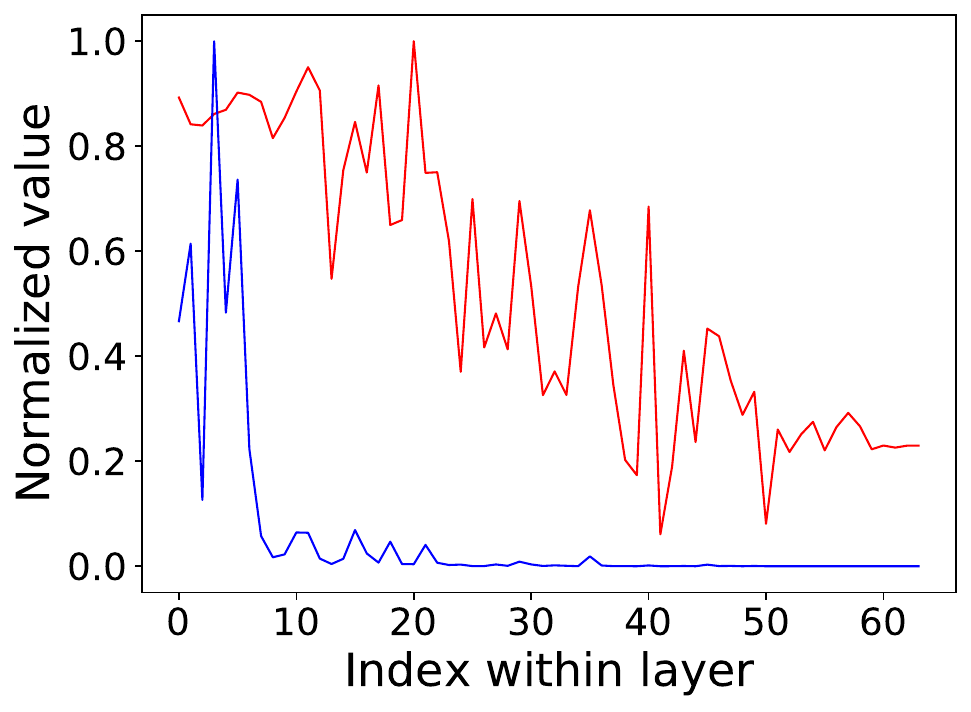}
    \end{minipage}
    \begin{minipage}[t]{0.19\linewidth}
        \centering{\phantom{aaaa}DenseNet-121 layer \#14}
        \centering
        \includegraphics[width=1\linewidth]{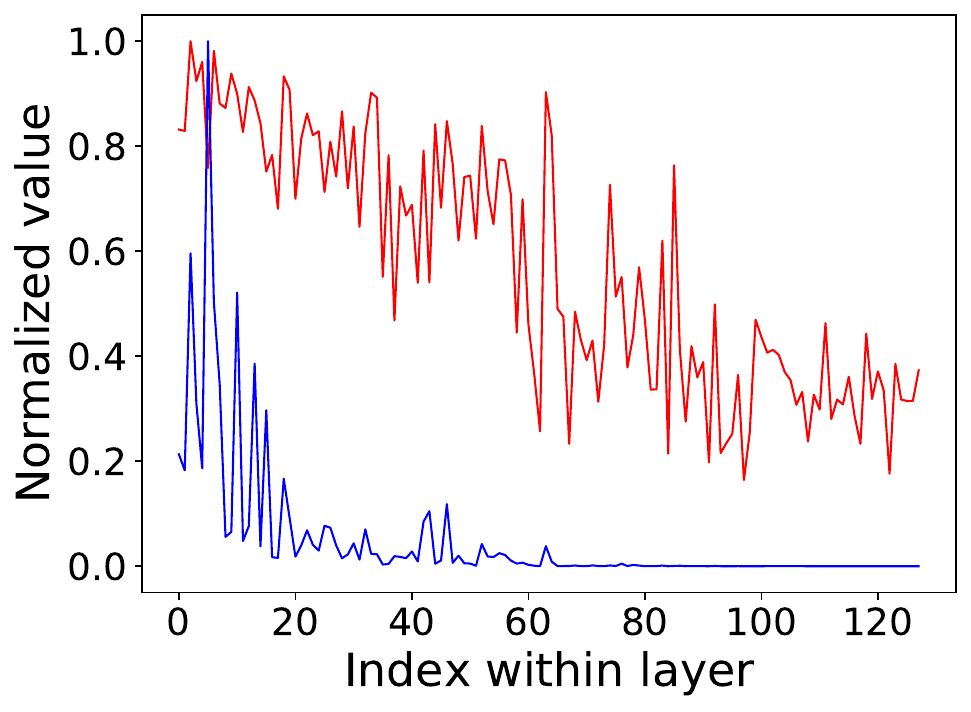}
    \end{minipage}
    \begin{minipage}[t]{0.19\linewidth}
        \centering{\phantom{aaaa}DenseNet-121 layer \#39}
        \centering
        \includegraphics[width=1\linewidth]{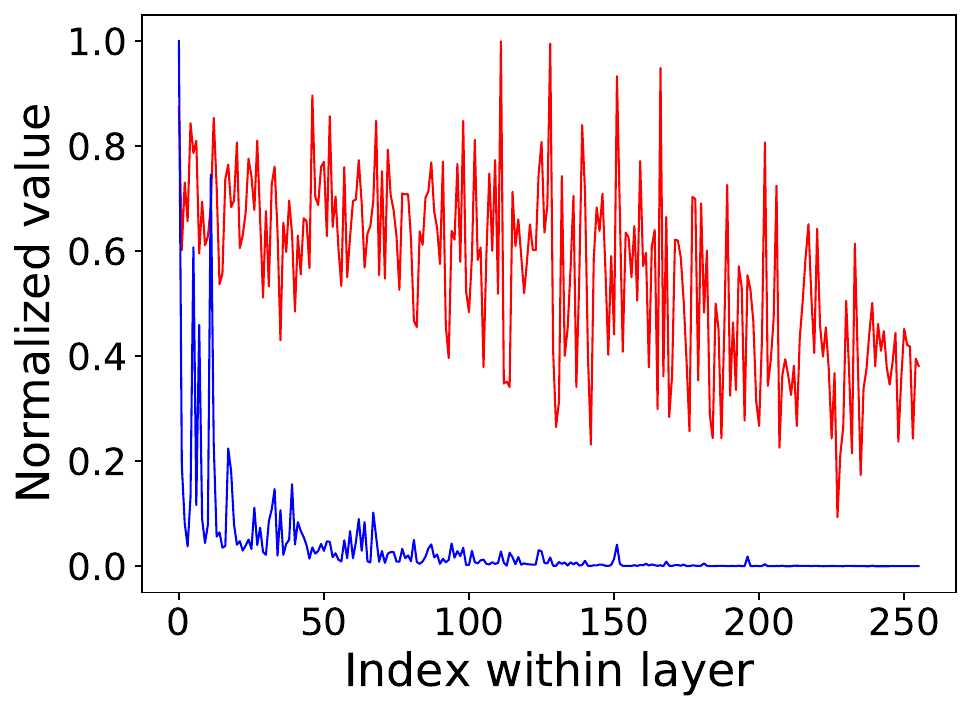}
    \end{minipage}
    \begin{minipage}[t]{0.19\linewidth}
        \centering{\phantom{aaaa}DenseNet-121 layer \#88}
        \centering
        \includegraphics[width=1\linewidth]{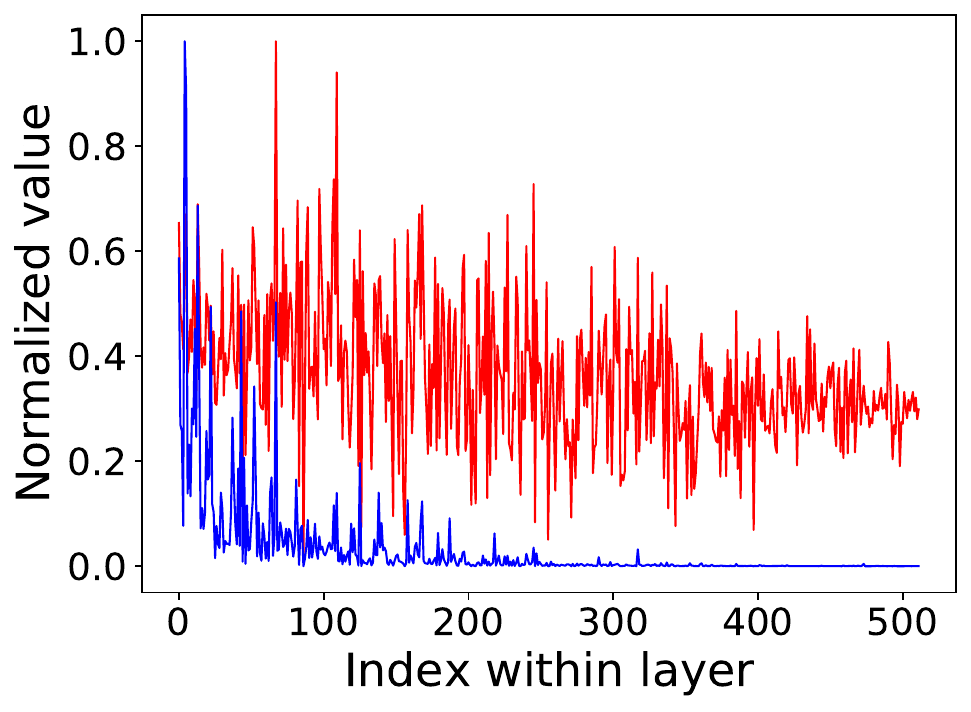}
    \end{minipage}
    \begin{minipage}[t]{0.19\linewidth}
        \centering{\phantom{aaaa}DenseNet-121 layer \#120}
        \centering
        \includegraphics[width=1\linewidth]{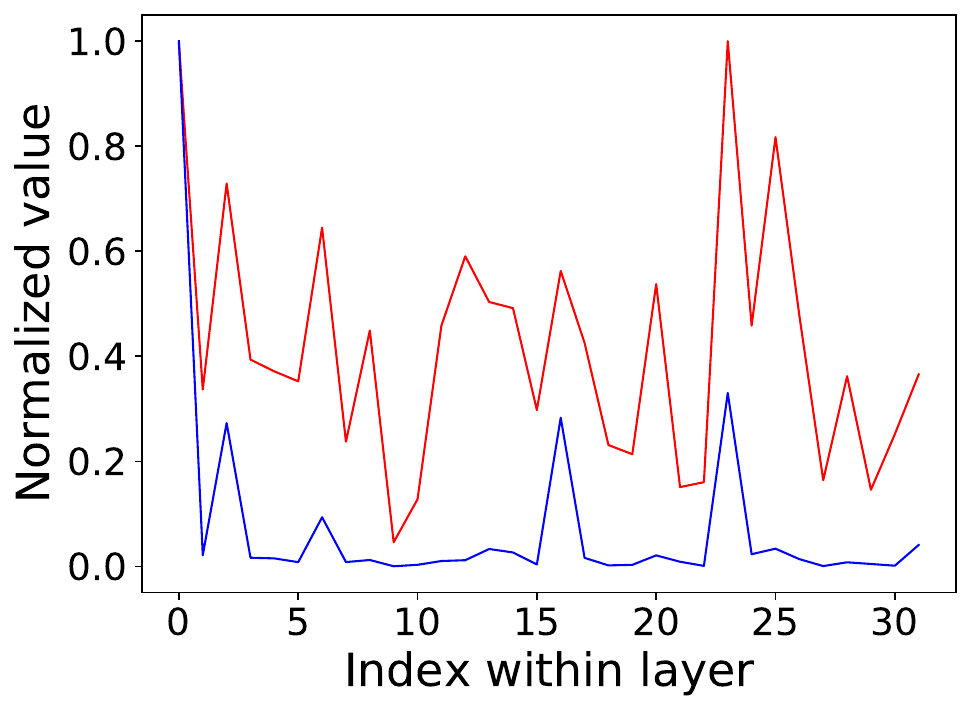}
    \end{minipage}
    \\
    \smallskip
    \begin{minipage}[t]{0.19\linewidth}
        \centering{\phantom{aaaa}ResNet-50 layer \#1}
        \centering
        \includegraphics[width=1\linewidth]{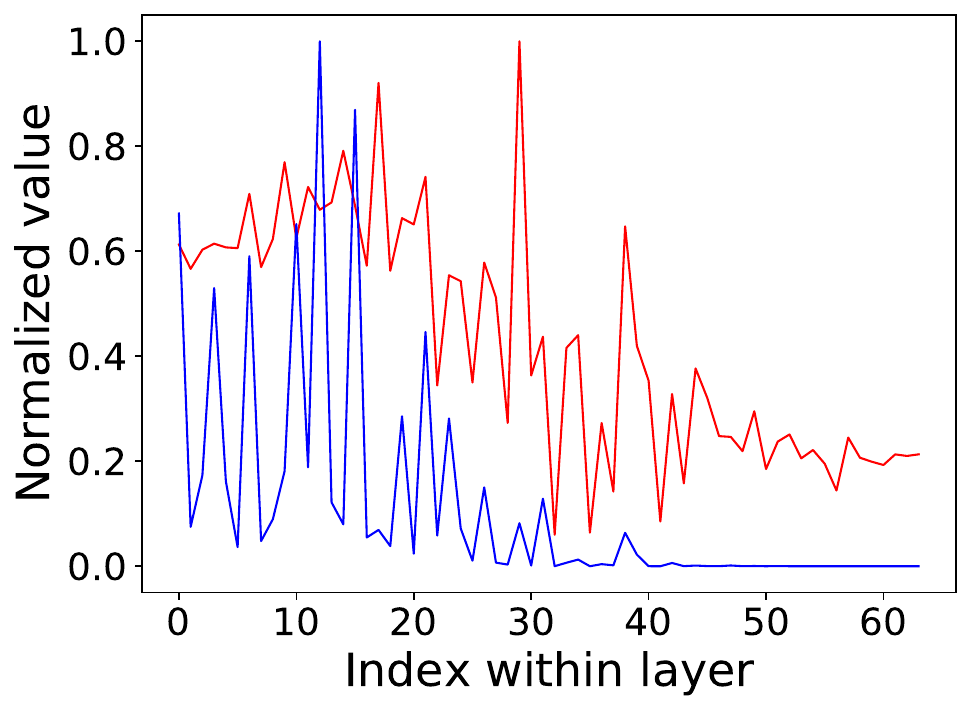}
    \end{minipage}
    \begin{minipage}[t]{0.19\linewidth}
        \centering{\phantom{aaaa}ResNet-50 layer \#11}
        \centering
        \includegraphics[width=1\linewidth]{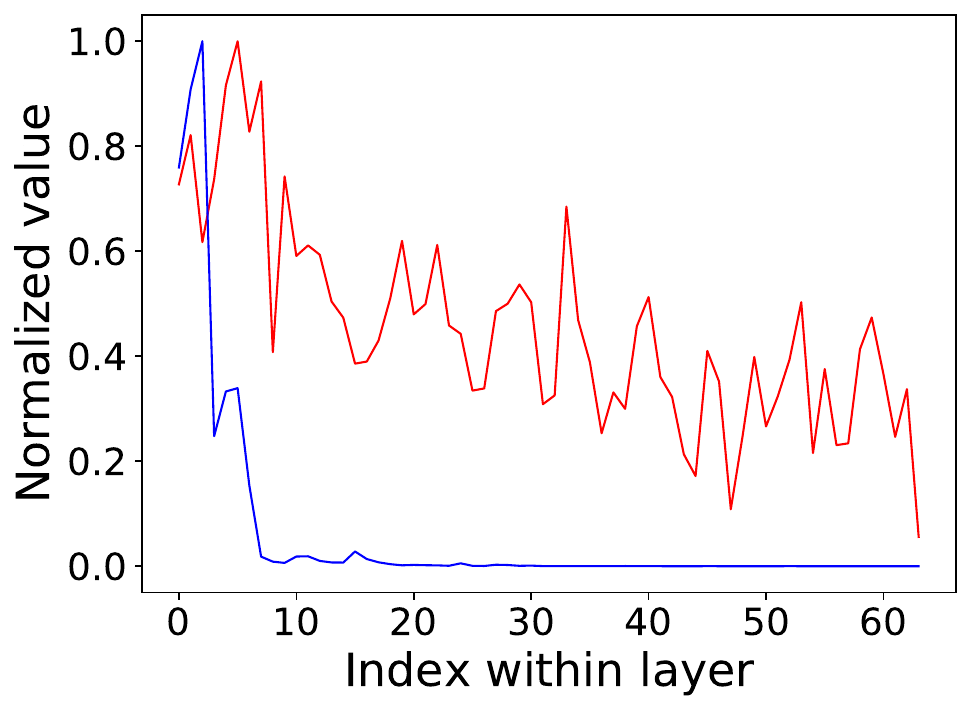}
    \end{minipage}
    \begin{minipage}[t]{0.19\linewidth}
        \centering{\phantom{aaaa}ResNet-50 layer \#24}
        \centering
        \includegraphics[width=1\linewidth]{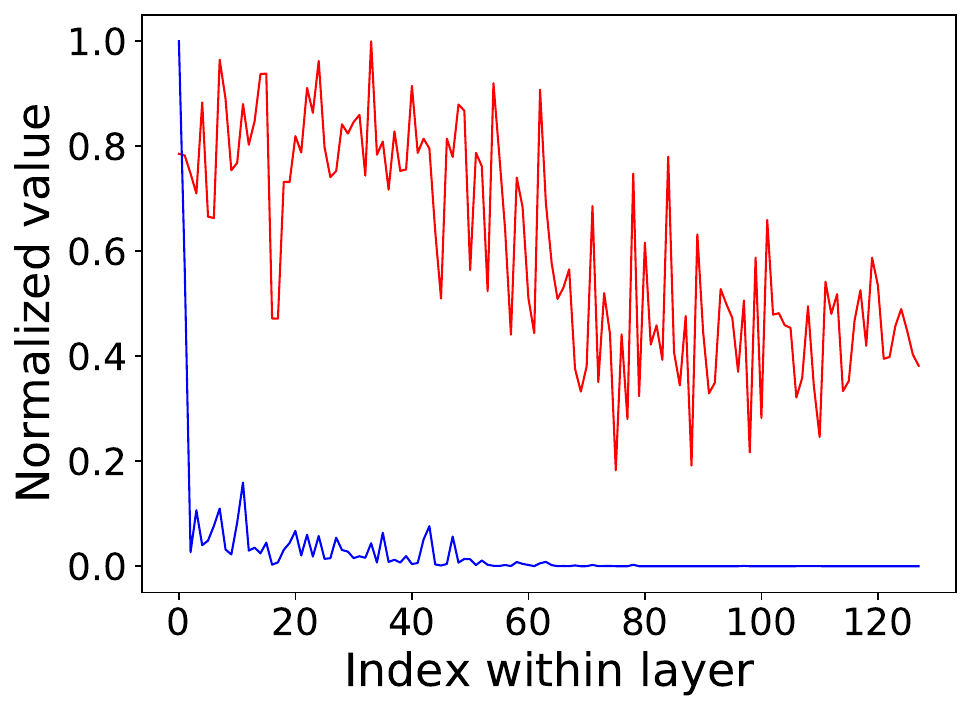}
    \end{minipage}
    \begin{minipage}[t]{0.19\linewidth}
        \centering{\phantom{aaaa}ResNet-50 layer \#43}
        \centering
        \includegraphics[width=1\linewidth]{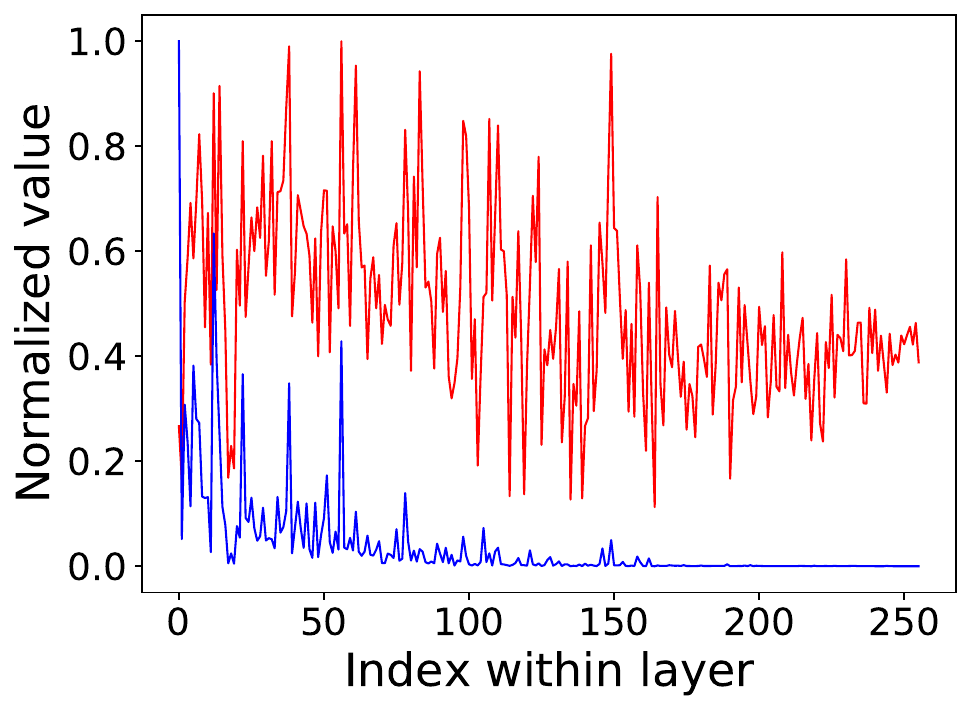}
    \end{minipage}
    \begin{minipage}[t]{0.19\linewidth}
        \centering{\phantom{aaaa}ResNet-50 layer \#53}
        \centering
        \includegraphics[width=1\linewidth]{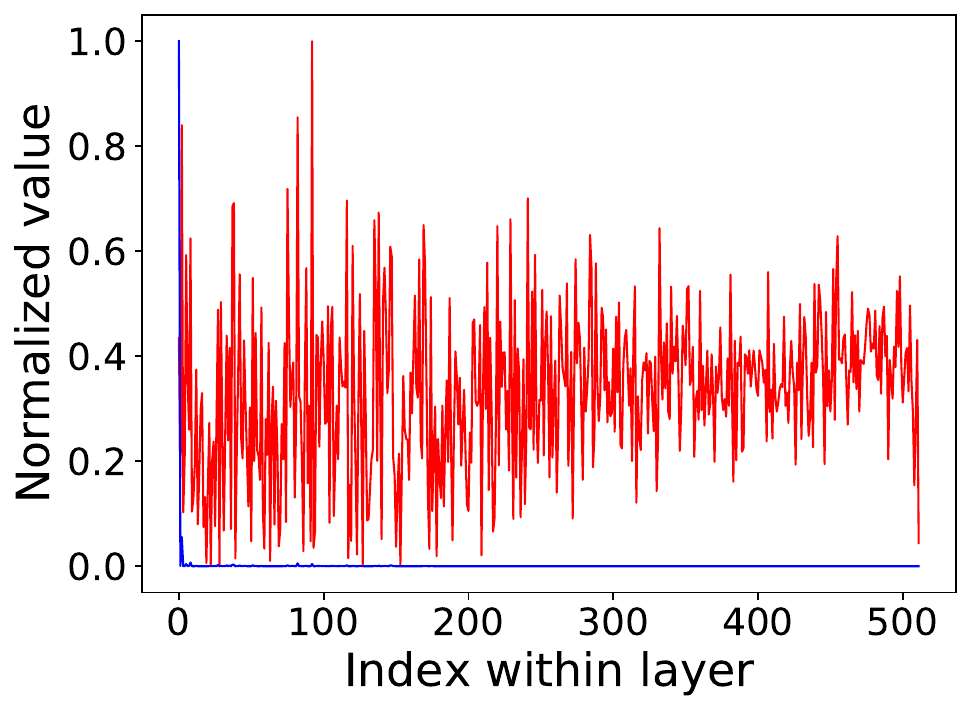}
    \end{minipage}
    \caption{The basis scaling factors (red) vs. the first-order Taylor approximation of importance (blue) of three different ImageNet pre-trained models on CIFAR-10. The scaling factors in each layer are arranged in descending order of the singular values. Note that ResNet-50 actually has 53 convolutional layers because of the four convolutional shortcuts.}
    \label{fig:scalars_taylorfo}
\end{figure*}

Given a pre-trained convolutional layer with non-trainable weights $\mathbf{W}$ and bias $\mathbf{b} \in \mathbb{R}^{c_o \times 1}$, we let $\mathbf{x}_i \in \mathbb{R}^{k \times 1}$ be a column vector of length $k = k_h \times k_w \times c_i$ which contains the features input to the convolutional layer at a spatial location. The output features $\mathbf{x}_o \in \mathbb{R}^{c_o \times 1}$ at the same spatial location can be obtained as:
\begin{align}
\label{eq:original_conv}
\mathbf{x}_o^\mathrm{T} = \mathbf{x}_i^\mathrm{T} \mathbf{W} + \mathbf{b}^\mathrm{T} = \mathbf{x}_i^\mathrm{T} \mathbf{U} \overline{\mathbf{V}}^\mathrm{T} + \mathbf{b}^\mathrm{T}
\end{align}
by using (\ref{eq:svd}) with $\overline{\mathbf{V}}^\mathrm{T} = \mathbf{\Sigma} \mathbf{V}^\mathrm{T}$. To rescale the basis vectors by their importance, we introduce a vector of basis scaling factors $\mathbf{s} \in \mathbb{R}^r$ of non-negative scalars, and (\ref{eq:original_conv}) is modified as:
\begin{align}
\label{eq:factorized_scaled_conv}
\mathbf{x}_o^\mathrm{T} = \mathbf{x}_i^\mathrm{T} \mathbf{U} \mathbf{S} \overline{\mathbf{V}}^\mathrm{T}  + \mathbf{b}^\mathrm{T}
\end{align}
with $\mathbf{S} \in \mathbb{R}^{r \times r}$ a diagonal matrix of $\mathbf{s}$. When $\mathbf{S} = \mathbf{I}$, (\ref{eq:original_conv}) and (\ref{eq:factorized_scaled_conv}) are identical. Using (\ref{eq:factorized_scaled_conv}), we can decompose the convolutional layer into two consecutive layers (Fig. \ref{fig:conv_decomposition}). The first layer is a regular convolutional layer with $\mathbf{U}$ as the convolutional weights and no bias. The second layer is the BasisScalingConv layer comprising $\mathbf{s}$, $\overline{\mathbf{V}}^\mathrm{T}$, and $\mathbf{b}$. $\overline{\mathbf{V}}^\mathrm{T}$ is used as the convolutional weights to transform the outputs from the previous layer back to the original space. Only $\mathbf{s}$ is trainable in the decomposed layers during transfer learning. When $\mathbf{s}$ is updated in each step (batch), each scalar in $\mathbf{s}$ rescales the corresponding row in $\overline{\mathbf{V}}^\mathrm{T}$. In fact, the scaling of the basis can be viewed as basis fine-tuning for improving accuracy. Instead of using (\ref{eq:factorized_scaled_conv}) as a single convolutional layer with $\mathbf{U} \mathbf{S} \overline{\mathbf{V}}^\mathrm{T}$ as the kernel, dividing into two layers reduces the number of weights and thus the computational complexity after the basis vectors are pruned.

\subsection{Basis Pruning with First-Order Taylor Approximation of Importance}
\label{sec:basis_pruning}

Our goal is to apply features trained from one dataset (e.g., ImageNet) to other datasets (e.g., CIFAR-10). Given a pre-trained model, we keep all layers up to the last convolutional layer and the associating BN and activation layers, and add a global average pooling layer and a final fully-connected layer for classification. For transfer learning with basis pruning, we first decompose every convolutional layer as presented in Section \ref{sec:layer_decomposition}. As BN layers are important for domain adaptation \cite{Workshop:Li:ICLRWorkshop2017:Revisiting}, they are trainable during transfer learning and are introduced after each convolutional layer if not present (e.g., VGGNet). Therefore, only the BN layers, the vector $\mathbf{s}$ in each BasisScalingConv layer, and the final fully-connected layer are trainable.

Although the magnitudes of the basis scaling factors can be used to indicate the relative importance of basis vectors, $l_1$ regularization is required to enhance sparsity for larger pruning ratios \cite{Conference:Liu:ICCV2017}. Finding the optimal $l_1$ regularization parameter that balances sparsity and accuracy is nontrivial and model dependent. To avoid this issue, we found that importance estimation by the first-order Taylor approximation (Taylor-FO) is a good alternative \cite{Conference:Molchanov:CVPR2019}. Taylor-FO measures the importance of a parameter by approximating the error induced by removing it. Using Taylor-FO, the importance score of a basis scaling factor $s$ can be approximated by:
\begin{align}
\label{eq:taylorFO}
\mathcal{I} = \left( gs \right)^2
\end{align}
with $g$ the gradient of the loss function with respect to $s$. Regardless of how importance scores are computed, they are normalized in each layer by the maximum importance score in that layer as some importance scores have different ranges in different layers. Fig. \ref{fig:scalars_taylorfo} shows that in the earlier layers, the basis scaling factors correspond to smaller singular values tend to be smaller, but the fluctuations become larger in the deeper layers and the correlation with the singular values is more difficult to observe. In contrast, the Taylor-FO importance scores are more correlated with the singular values, though they are different from the singular values to reflect the data-dependent importance.

\begin{figure}[t]
    \footnotesize
    \centering
    \begin{minipage}[t]{0.45\linewidth}
      \centering
      \includegraphics[width=1\linewidth]{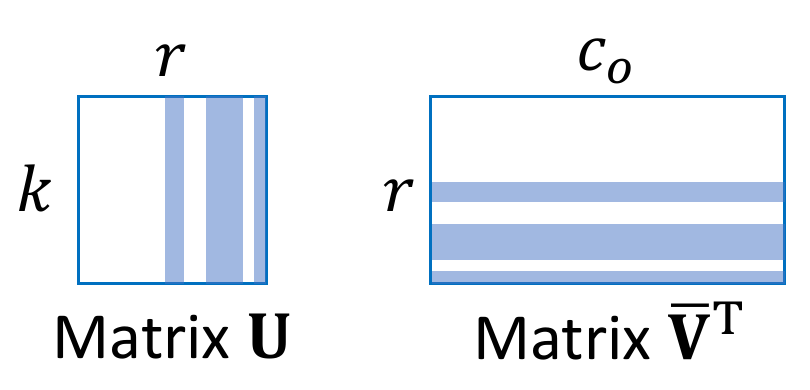} \\
      \centering{(a) Basis pruning.}
    \end{minipage}
    \hfill
    \vrule\
    \hfill
    \begin{minipage}[t]{0.45\linewidth}
      \centering
      \includegraphics[width=1\linewidth]{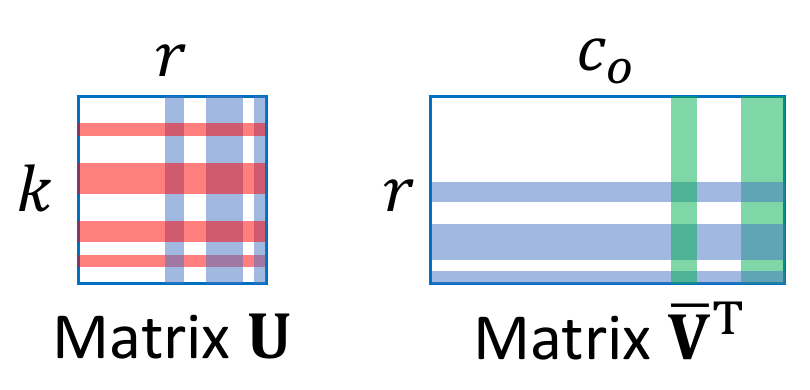} \\
      \centering{(b) Double pruning.}
    \end{minipage}
    \caption{Basis pruning and double pruning. (a) Pruning by removing the same number of basis vectors (blue) from $\mathbf{U}$ and $\overline{\mathbf{V}}^\mathrm{T}$. (b) With also input pruning (red) and output pruning (green).}
    \label{fig:double_prune}
\end{figure}

After training with enough epochs for the desired classification accuracy, the less important basis vectors are pruned (Fig. \ref{fig:double_prune}(a)). Let $r_p < r$ be the number of scaling factors that remain after pruning, then $\mathbf{U}$, $\mathbf{S}$, and $\overline{\mathbf{V}}^\mathrm{T}$ in (\ref{eq:factorized_scaled_conv}) become $\mathbf{U}_p \in \mathbb{R}^{k \times r_p}$, $\mathbf{S}_p \in \mathbb{R}^{r_p \times r_p}$, and $\overline{\mathbf{V}}^\mathrm{T}_p \in \mathbb{R}^{r_p \times c_o}$, respectively. As the sizes of $\mathbf{x}_i$, $\mathbf{x}_o$, and $\mathbf{b}$ are unaltered, basis pruning only affects the convolutional layer being pruned but not the subsequent layer. Therefore, basis pruning can be applied to virtually all network architectures. In contrast, pruning in the original space requires pruning of the subsequent convolutional layer, which can be complicated with skip connections.

\subsection{Double Pruning}
\label{sec:double_prune}

In basis pruning, only the output channels of $\mathbf{U}$ and the input channels of $\overline{\mathbf{V}}^\mathrm{T}$ are pruned (Fig. \ref{fig:double_prune}(a)). With this unique characteristic, larger pruning ratios can be achieved by pruning also the input channels of $\mathbf{U}$ and the output channels of $\overline{\mathbf{V}}^\mathrm{T}$ (Fig. \ref{fig:double_prune}(b)). This can be achieved by pruning the output channels of $\overline{\mathbf{V}}^\mathrm{T}$ as the subsequent input channels of $\mathbf{U}$ are pruned accordingly. In fact, after basis pruning, the BasisScalingConv layers can be treated as regular convolutional layers and any existing pruning algorithms can be applied.

\subsection{Pruning Procedure}
\label{sec:prune_procedure}

\begin{enumerate}[noitemsep]
  \item Given a pre-trained model, keep all layers up to the last convolutional layer and the associating BN and activation layers. Add a global average pooling layer and a final fully-connected layer for classification. Insert BN layers if needed.
  \item \label{proc:decompose} Decompose each convolutional layer into a convolutional layer and a BasisScalingConv layer (Section \ref{sec:layer_decomposition}).
  \item \label{proc:first_train} Train the model with only the BN layers, the basis scaling factors, and the final fully-connected layer trainable with the desired number of epochs.
  \item \label{proc:basis_prune} Perform basis pruning by removing the basis vectors whose normalized importance scores are lower than a given threshold (Section \ref{sec:basis_pruning}). Train the pruned model as in Step \ref{proc:first_train}.
  \item \label{proc:double_prune} Perform pruning by removing the output channels of the BasisScalingConv layers whose normalized importance scores are lower than a given threshold  (Section \ref{sec:double_prune}). Train the pruned model as in Step \ref{proc:first_train}.
\end{enumerate}
The pseudocode is in Algorithm \ref{alg:prune}. Multiple iterations can be performed in Step \ref{proc:basis_prune} and \ref{proc:double_prune}, though we found that one iteration is sufficient. In Step \ref{proc:first_train}, each scaling factor in the BasisScalingConv or BN layers modifies the weights of a basis vector or a filter as a whole. We regard this as basis-based or filter-based fine-tuning which is more efficient than fine-tuning individual weights.


To determine the thresholds in Step \ref{proc:basis_prune} and \ref{proc:double_prune}, we decide the percentage of basis vectors or filters to be removed from the entire model and compute the corresponding threshold. This can be done as the importance scores are normalized (Section \ref{sec:basis_pruning}). To minimize manual involvements and improve training efficiency, different from \cite{Conference:Li:ICLR2017:Pruning}, no manual layer selections are performed. Moreover, we do not use gradual filters removal \cite{Conference:Molchanov:CVPR2019} or filter-by-filter pruning \cite{Conference:Lin:CVPR2020}. Only the percentage of basis vectors or filters to be removed is needed.

\begin{algorithm}[t]
\scriptsize
\caption{The proposed pruning algorithm.}\label{alg:prune}
\begin{algorithmic}[1]
\Require A pre-trained model with a set of convolutional layers, each with weights $\mathbf{W}$ and bias $\mathbf{b}$.
\Require A training dataset and a validation dataset of the target domain.
\item[]
\Function{finetuning}{$model$} \Comment{For multiple calls}
\State Train $model$ using the training dataset with only the BN layers, the basis scaling factors, and the final fully-connected layer trainable with the desired number of epochs.
\State \Return A fine-tuned model.
\EndFunction
\item[]
\item[\textbf{Convolutional layer decomposition}]
\State Keep all layers up to the last convolutional layer and the associating BN and activation layers with the convolutional weights frozen. Add a global average pooling layer and a final fully-connected layer for classification. Insert BN layers if needed.
\ForEach{convolutional layer} \Comment{Layer decomposition}
    \State Use the compact SVD to compute $\mathbf{W} = \mathbf{U} \overline{\mathbf{V}}^\mathrm{T}$ in (\ref{eq:original_conv}).
    \State Replace the convolutional layer by two consecutive layers: 1) a convolutional layer with weights $\mathbf{U}$ and no bias; 2) a BasisScalingConv layer with weights $\overline{\mathbf{V}}^\mathrm{T}$, bias $\mathbf{b}$, and basis scaling factors $\mathbf{s}$ in (\ref{eq:factorized_scaled_conv}).
\EndFor
\State \Call{finetuning}{$model$}
\item[]
\item[\textbf{Basis pruning -- remove basis vectors}]
\State Compute the importance scores of basis scaling factors $\mathbf{s}$ using (\ref{eq:taylorFO}) with the validation dataset. The scores are normalized for each layer.
\State Compute the threshold of the importance scores based on the desired percentage of basis vectors to be removed.
\State Remove basis vectors with importance scores lower than the threshold.
\State \Call{finetuning}{$model$}
\item[]
\item[\textbf{Double pruning -- remove filters in BasisScalingConv layers}]
\State Compute the importance scores of filters in BasisScalingConv layers using any existing method with the validation dataset. The scores are normalized for each layer.
\State Compute the threshold of the importance scores based on the desired percentage of filters to be removed.
\State Remove filters with importance scores lower than the threshold.
\State \Call{finetuning}{$model$}
\end{algorithmic}
\end{algorithm}

\section{Experiments}

\subsection{Models and Datasets}

To study the characteristics of our framework, we performed transfer learning with three ImageNet \cite{Conference:Deng:CVPR2009} pre-trained models on three other datasets. ImageNet was used as the source dataset because of its abundant features trained from 1.2 million images. The models studied were VGG-16 \cite{Conference:Simonyan:ICLR2015:Very}, DenseNet-121 \cite{Conference:Huang:CVPR2017}, and ResNet-50 \cite{Conference:He:CVPR2016}, which are commonly used for transfer learning. The three datasets include CIFAR-10 \cite{Report:Krizhevsky:CIFAR2009}, MNIST \cite{Journal:Lecun:IEEEProceedings1998}, and Fashion-MNIST \cite{Journal:Xiao:arXiv2017}. The CIFAR-10 dataset consists of 32$\times$32 color images in 10 classes of animals and vehicles, with 50k training images and 10k test images. The MNIST dataset of handwritten digits (0 to 9) has 60k 28$\times$28 grayscale training images and 10k test images in 10 classes. The Fashion-MNIST dataset has a training set of 60k 28$\times$28 grayscale training images and 10k test images in 10 classes of fashion categories, which can be used as a drop-in replacement for MNIST. Each set of training images was split into 90\% for training and 10\% for validation. The results on the test images are reported.

\subsection{Tested Frameworks}

Given a pre-trained model, we kept all layers up to the last convolutional layer and the associating BN and activation layers, and added a global average pooling layer and a final fully-connected layer. The BN layers and the final fully-connected layer were trainable while others were frozen. Different frameworks were tested using this configuration:
\begin{itemize}[noitemsep]
  \item \textbf{Baseline}: no pruning.
  \item \textbf{L1} \cite{Conference:Li:ICLR2017:Pruning}: pruning using the L1 norms of filters as the importance scores.
  \item \textbf{Taylor-FO} \cite{Conference:Molchanov:CVPR2019}: pruning by the Taylor-FO importance scores with the gradients computed from the validation dataset.
  \item \textbf{HRank} \cite{Conference:Lin:CVPR2020}: pruning by the HRank importance scores which are the matrix ranks of the feature maps of the validation dataset.
  \item \textbf{Basis}: all convolutional layers were decomposed. Pruning by the Taylor-FO importance scores of the basis scaling factors (Section \ref{sec:basis_pruning}).
  \item \textbf{Basis + Taylor-FO}: double pruning by the Taylor-FO importance scores (Section \ref{sec:double_prune}).
  \item \textbf{Basis + HRank}: double pruning by the HRank importance scores (Section \ref{sec:double_prune}).
\end{itemize}

For the frameworks without layer decompositions, the pruning procedure in Section \ref{sec:prune_procedure} was applied without Step \ref{proc:decompose} and \ref{proc:basis_prune}, and the targeting layers in Step \ref{proc:double_prune} became the convolutional layers. For all frameworks, only one iteration was performed in Step \ref{proc:basis_prune} and \ref{proc:double_prune}. As the L1 framework was less effective in our experiments, we did not use it for double pruning. Note that our goal is to study the pruning capabilities of our frameworks but not competing for the best accuracy on specific datasets.

\subsection{Training Strategy}

For transfer learning, as the network architectures of the ImageNet pre-trained models were created for the image size of 224$\times$224, directly applying them to the target datasets of much smaller image sizes leads to insufficient spatial sizes (e.g., feature maps of size 1$\times$1) in the deeper layers and thus poor performance. Therefore, we enlarged the image sizes by four times in each dimension, i.e., 128$\times$128 for CIFAR-10 and 112$\times$112 for MNIST and Fashion-MNIST. Image augmentation was used to reduce overfitting, with $\pm$15\% of shifting in height and width for all datasets, random horizontal flipping for CIFAR-10 and Fashion-MNIST, and $\pm$15$^\circ$ of rotation for MNIST. Every image was zero-centered in intensity. Dropout with rate 0.5 was applied before the final fully connected layer. Stochastic gradient descent (SGD) with cosine annealing \cite{Conference:Loshchilov:ICLR2017:SGDR} was used as the learning rate scheduler without restarts, with the maximum and minimum learning rates as $10^{-1}$ and $10^{-4}$, respectively. The SGD optimizer was used with momentum of 0.9 and a batch size of 128. There were 100 epochs for each training. All scaling factors were initialized to 0.5 and constrained to be non-negative.

The implementation was in Keras with TensorFlow backend. Each experiment was performed on an NVIDIA Tesla V100 GPU with 16 GB memory.

\begin{table}[t]
\caption{Pruning results on CIFAR-10 with ImageNet pre-trained models. PR = pruning ratio. Proposed frameworks are in bold and the best results are in blue. Results that were worse than all proposed frameworks are in red. The images were upsampled to 128$\times$128.}
\label{table:cifar-10}

\newcolumntype{R}{>{\raggedleft\arraybackslash}X}
\scriptsize

\begin{tabular*}{\columnwidth}{@{\extracolsep{\fill}}lcr@{\hskip 0.5ex}rr@{\hskip 0.5ex}r}
\toprule
\multicolumn{1}{c}{Framework} & \multicolumn{1}{c}{Accuracy} & \multicolumn{2}{c}{Parameters (PR)} & \multicolumn{2}{c}{FLOPs (PR)} \\
\midrule
VGG-16 & 90.9\% & 14.74M & (0.0\%) & 5.03G & (0.0\%) \\
L1 \cite{Conference:Li:ICLR2017:Pruning} & 88.0\% & {\color{red}\textbf{13.42M}} & {\color{red}\textbf{(8.9\%)}} & 3.15G & (37.5\%) \\
Taylor-FO \cite{Conference:Molchanov:CVPR2019} & 90.0\% & 6.57M & (55.4\%) & 2.48G & (50.8\%) \\
HRank \cite{Conference:Lin:CVPR2020} & 90.9\% & {\color{red}\textbf{11.67M}} & {\color{red}\textbf{(20.8\%)}} & {\color{red}\textbf{4.11G}} & {\color{red}\textbf{(18.4\%)}} \\
\textbf{Basis} & 91.2\% & 7.99M & (45.8\%) & 3.21G & (36.2\%) \\
\textbf{Basis + Taylor-FO} & 90.5\% & {\color{blue}\textbf{5.42M}} & {\color{blue}\textbf{(63.2\%)}} & {\color{blue}\textbf{2.45G}} & {\color{blue}\textbf{(51.2\%)}} \\
\textbf{Basis + HRank} & 90.9\% & 7.13M & (51.6\%) & 2.95G & (41.3\%) \\
\midrule
DenseNet-121 & 95.2\% & 7.05M & (0.0\%) & 0.93G & (0.0\%) \\
L1 \cite{Conference:Li:ICLR2017:Pruning} & 93.6\% & {\color{red}\textbf{6.02M}} & {\color{red}\textbf{(14.5\%)}} & {\color{red}\textbf{0.70G}} & {\color{red}\textbf{(24.7\%)}} \\
Taylor-FO \cite{Conference:Molchanov:CVPR2019} & 95.1\% & {\color{blue}\textbf{4.15M}} & {\color{blue}\textbf{(41.2\%)}} & {\color{blue}\textbf{0.57G}} & {\color{blue}\textbf{(38.8\%)}} \\
HRank \cite{Conference:Lin:CVPR2020} & 94.3\% & {\color{red}\textbf{5.41M}} & {\color{red}\textbf{(23.2\%)}} & {\color{red}\textbf{0.78G}} & {\color{red}\textbf{(15.9\%)}} \\
\textbf{Basis} & 94.4\% & 4.75M & (32.6\%) & 0.65G & (29.9\%) \\
\textbf{Basis + Taylor-FO} & 94.7\% & 4.43M & (37.2\%) & 0.60G & (35.7\%) \\
\textbf{Basis + HRank} & 94.1\% & 4.20M & (40.4\%) & 0.60G & (35.5\%) \\
\midrule
ResNet-50 & 92.4\% & 23.61M & (0.0\%) & 1.29G & (0.0\%) \\
L1 \cite{Conference:Li:ICLR2017:Pruning} & 92.1\% & {\color{red}\textbf{23.15M}} & {\color{red}\textbf{(1.9\%)}} & {\color{red}\textbf{1.18G}} & {\color{red}\textbf{(8.3\%)}} \\
Taylor-FO \cite{Conference:Molchanov:CVPR2019} & 91.4\% & {\color{red}\textbf{10.89M}} & {\color{red}\textbf{(53.9\%)}} & {\color{red}\textbf{0.75G}} & {\color{red}\textbf{(42.0\%)}} \\
HRank \cite{Conference:Lin:CVPR2020} & 91.8\% & {\color{red}\textbf{17.67M}} & {\color{red}\textbf{(25.2\%)}} & {\color{red}\textbf{1.00G}} & {\color{red}\textbf{(22.5\%)}} \\
\textbf{Basis} & 92.2\% & 7.14M & (69.7\%) & 0.61G & (52.6\%) \\
\textbf{Basis + Taylor-FO} & 91.6\% & {\color{blue}\textbf{6.00M}} & {\color{blue}\textbf{(74.6\%)}} & {\color{blue}\textbf{0.52G}} & {\color{blue}\textbf{(59.8\%)}} \\
\textbf{Basis + HRank} & 92.1\% & 6.39M & (72.9\%) & 0.55G & (57.0\%) \\
\bottomrule
\end{tabular*}
\end{table}

\begin{table}[t]
\caption{Pruning results on MNIST with ImageNet pre-trained models. PR = pruning ratio. Proposed frameworks are in bold and the best results are in blue. Results that were worse than all proposed frameworks are in red. The images were upsampled to 112$\times$112.}
\label{table:mnist}

\newcolumntype{R}{>{\raggedleft\arraybackslash}X}
\scriptsize

\begin{tabular*}{\columnwidth}{@{\extracolsep{\fill}}lcr@{\hskip 0.5ex}rr@{\hskip 0.5ex}r}
\toprule
\multicolumn{1}{c}{Framework} & \multicolumn{1}{c}{Accuracy} & \multicolumn{2}{c}{Parameters (PR)} & \multicolumn{2}{c}{FLOPs (PR)} \\
\midrule
VGG-16 & 99.5\% & 14.74M & (0.0\%) & 3.85G & (0.0\%) \\
L1 \cite{Conference:Li:ICLR2017:Pruning} & 99.3\% & {\color{red}\textbf{4.45M}} & {\color{red}\textbf{(69.8\%)}} & {\color{red}\textbf{0.55G}} & {\color{red}\textbf{(85.8\%)}} \\
Taylor-FO \cite{Conference:Molchanov:CVPR2019} & 99.3\% & {\color{red}\textbf{0.73M}} & {\color{red}\textbf{(95.1\%)}} & 0.29G & (92.4\%) \\
HRank \cite{Conference:Lin:CVPR2020} & 99.4\% & {\color{red}\textbf{3.71M}} & {\color{red}\textbf{(74.8\%)}} & {\color{red}\textbf{1.51G}} & {\color{red}\textbf{(60.8\%)}} \\
\textbf{Basis} & 99.4\% & 0.64M & (95.6\%) & 0.35G & (90.8\%) \\
\textbf{Basis + Taylor-FO} & 99.3\% & {\color{blue}\textbf{0.16M}} & {\color{blue}\textbf{(98.9\%)}} & {\color{blue}\textbf{0.12G}} & {\color{blue}\textbf{(97.0\%)}} \\
\textbf{Basis + HRank} & 99.3\% & 0.35M & (97.6\%) & 0.25G & (93.5\%) \\
\midrule
DenseNet-121 & 99.6\% & 7.05M & (0.0\%) & 0.71G & (0.0\%) \\
L1 \cite{Conference:Li:ICLR2017:Pruning} & 99.2\% & {\color{blue}\textbf{0.11M}} & {\color{blue}\textbf{(98.5\%)}} & {\color{blue}\textbf{0.01G}} & {\color{blue}\textbf{(98.6\%)}} \\
Taylor-FO \cite{Conference:Molchanov:CVPR2019} & 99.3\% & 0.12M & (98.2\%) & 0.01G & (98.3\%) \\
HRank \cite{Conference:Lin:CVPR2020} & 99.1\% & 0.30M & (95.8\%) & {\color{red}\textbf{0.11G}} & {\color{red}\textbf{(83.8\%)}} \\
\textbf{Basis} & 99.1\% & 0.37M & (94.8\%) & 0.03G & (96.1\%) \\
\textbf{Basis + Taylor-FO} & 99.4\% & 0.19M & (97.4\%) & 0.01G & (98.3\%) \\
\textbf{Basis + HRank} & 99.0\% & 0.32M & (95.4\%) & 0.02G & (96.7\%) \\
\midrule
ResNet-50 & 99.4\% & 23.61M & (0.0\%) & 1.05G & (0.0\%) \\
L1 \cite{Conference:Li:ICLR2017:Pruning} & 99.0\% & {\color{red}\textbf{6.14M}} & {\color{red}\textbf{(74.0\%)}} & {\color{red}\textbf{0.24G}} & {\color{red}\textbf{(76.9\%)}} \\
Taylor-FO \cite{Conference:Molchanov:CVPR2019} & 99.2\% & {\color{red}\textbf{5.74M}} & {\color{red}\textbf{(75.7\%)}} & {\color{red}\textbf{0.23G}} & {\color{red}\textbf{(77.7\%)}} \\
HRank \cite{Conference:Lin:CVPR2020} & 99.3\% & {\color{red}\textbf{6.54M}} & {\color{red}\textbf{(72.3\%)}} & {\color{red}\textbf{0.44G}} & {\color{red}\textbf{(57.5\%)}} \\
\textbf{Basis} & 99.1\% & 0.55M & (97.7\%) & 0.05G & (95.1\%) \\
\textbf{Basis + Taylor-FO} & 99.1\% & {\color{blue}\textbf{0.48M}} & {\color{blue}\textbf{(98.0\%)}} & {\color{blue}\textbf{0.04G}} & {\color{blue}\textbf{(95.9\%)}} \\
\textbf{Basis + HRank} & 99.0\% & 0.50M & (97.9\%) & 0.04G & (95.7\%) \\
\bottomrule
\end{tabular*}
\end{table}

\begin{table}[t]
\caption{Pruning results on Fashion-MNIST with ImageNet pre-trained models. PR = pruning ratio. Proposed frameworks are in bold and the best results are in blue. Results that were worse than all proposed frameworks are in red. The images were upsampled to 112$\times$112.}
\label{table:fashion-mnist}

\newcolumntype{R}{>{\raggedleft\arraybackslash}X}
\scriptsize

\begin{tabular*}{\columnwidth}{@{\extracolsep{\fill}}lcr@{\hskip 0.5ex}rr@{\hskip 0.5ex}r}
\toprule
\multicolumn{1}{c}{Framework} & \multicolumn{1}{c}{Accuracy} & \multicolumn{2}{c}{Parameters (PR)} & \multicolumn{2}{c}{FLOPs (PR)} \\
\midrule
VGG-16 & 93.2\% & 14.74M & (0.0\%) & 3.85G & (0.0\%) \\
L1 \cite{Conference:Li:ICLR2017:Pruning} & 92.9\% & {\color{red}\textbf{13.42M}} & {\color{red}\textbf{(8.9\%)}} & {\color{red}\textbf{2.41G}} & {\color{red}\textbf{(37.4\%)}} \\
Taylor-FO \cite{Conference:Molchanov:CVPR2019} & 92.5\% & {\color{red}\textbf{2.64M}} & {\color{red}\textbf{(82.1\%)}} & {\color{red}\textbf{1.04G}} & {\color{red}\textbf{(73.1\%)}} \\
HRank \cite{Conference:Lin:CVPR2020} & 92.2\% & {\color{red}\textbf{3.54M}} & {\color{red}\textbf{(76.0\%)}} & {\color{red}\textbf{1.66G}} & {\color{red}\textbf{(56.9\%)}} \\
\textbf{Basis} & 92.3\% & 1.37M & (90.7\%) & 0.68G & (82.3\%) \\
\textbf{Basis + Taylor-FO} & 92.4\% & {\color{blue}\textbf{0.90M}} & {\color{blue}\textbf{(93.9\%)}} & {\color{blue}\textbf{0.51G}} & {\color{blue}\textbf{(86.6\%)}} \\
\textbf{Basis + HRank} & 92.0\% & 0.98M & (93.4\%) & 0.54G & (85.9\%) \\
\midrule
DenseNet-121 & 93.7\% & 7.05M & (0.0\%) & 0.71G & (0.0\%) \\
L1 \cite{Conference:Li:ICLR2017:Pruning} & 93.1\% & {\color{red}\textbf{2.54M}} & {\color{red}\textbf{(63.9\%)}} & {\color{red}\textbf{0.17G}} & {\color{red}\textbf{(76.6\%)}} \\
Taylor-FO \cite{Conference:Molchanov:CVPR2019} & 92.8\% & 0.94M & (86.6\%) & 0.11G & (85.0\%) \\
HRank \cite{Conference:Lin:CVPR2020} & 92.7\% & 1.06M & (85.0\%) & {\color{red}\textbf{0.28G}} & {\color{red}\textbf{(60.0\%)}} \\
\textbf{Basis} & 93.4\% & 1.13M & (84.0\%) & 0.14G & (80.0\%) \\
\textbf{Basis + Taylor-FO} & 92.7\% & {\color{blue}\textbf{0.63M}} & {\color{blue}\textbf{(91.0\%)}} & {\color{blue}\textbf{0.07G}} & {\color{blue}\textbf{(90.6\%)}} \\
\textbf{Basis + HRank} & 93.5\% & 0.83M & (88.3\%) & 0.11G & (85.1\%) \\
\midrule
ResNet-50 & 94.0\% & 23.61M & (0.0\%) & 1.05G & (0.0\%) \\
L1 \cite{Conference:Li:ICLR2017:Pruning} & 93.5\% & {\color{red}\textbf{22.04M}} & {\color{red}\textbf{(6.6\%)}} & {\color{red}\textbf{0.86G}} & {\color{red}\textbf{(18.1\%)}} \\
Taylor-FO \cite{Conference:Molchanov:CVPR2019} & 93.0\% & {\color{red}\textbf{9.12M}} & {\color{red}\textbf{(61.4\%)}} & {\color{red}\textbf{0.47G}} & {\color{red}\textbf{(55.1\%)}} \\
HRank \cite{Conference:Lin:CVPR2020} & 93.2\% & {\color{red}\textbf{4.73M}} & {\color{red}\textbf{(80.0\%)}} & {\color{red}\textbf{0.28G}} & {\color{red}\textbf{(73.4\%)}} \\
\textbf{Basis} & 93.3\% & 1.99M & (91.6\%) & 0.21G & (79.6\%) \\
\textbf{Basis + Taylor-FO} & 93.4\% & 1.47M & (93.8\%) & 0.16G & (85.1\%) \\
\textbf{Basis + HRank} & 93.4\% & {\color{blue}\textbf{1.46M}} & {\color{blue}\textbf{(93.8\%)}} & {\color{blue}\textbf{0.15G}} & {\color{blue}\textbf{(85.3\%)}} \\
\bottomrule
\end{tabular*}
\end{table}

\subsection{Pruning Performance on Transfer Learning}

Table \ref{table:cifar-10}, \ref{table:mnist}, \ref{table:fashion-mnist} show the comparisons among frameworks. To compare the pruning capabilities, we obtained the largest pruning ratios while keeping the accuracy reductions less than 1\%. In a few occasions, this 1\% requirement could not be achieved even with only 10\% of channels removal. For MNIST, we kept the accuracies $\geq 99\%$.

The pruning ratios were inversely proportional to the difficulties of the classification problems. The largest parameter pruning ratios of CIFAR-10, MNIST, and Fashion-MNIST were 74.6\%, 98.9\%, and 93.9\%, respectively, all achieved by our Basis + Taylor-FO framework. For ResNet-50, our frameworks outperformed the others by a large margin on all datasets. This is because basis pruning can be applied to all convolutional layers regardless of the existence of skip connections. Our frameworks were also dominant on VGG-16. Although they were less effective with DenseNet-121 on CIFAR-10 and MNIST, our best results were only lower than the best ones by less than 1.1\% in parameters and 3.1\% in FLOPs.

In seven out of the nine combinations (3 architectures $\times$ 3 datasets), our double pruning frameworks outperformed the others. This was often (6 out of 7) achieved by Basis + Taylor-FO and once by Basis + HRank.

\subsection{Trainable Model Parameters}

Table \ref{table:params} shows the numbers of total and trainable model parameters after layer decomposition and before pruning. Although the total numbers of parameters became larger because of layer decomposition, the trainable parameters were less than 1.2\% and could be as low as 0.1\% of the total numbers of parameters. Therefore, our framework is advantageous for transfer learning with limited data.

\begin{table}[t]
\caption{Numbers of total and trainable parameters before pruning.}
\label{table:params}

\newcolumntype{R}{>{\raggedleft\arraybackslash}X}
\scriptsize

\begin{tabular*}{\columnwidth}{@{\extracolsep{\fill}}lrr}
\toprule
\multicolumn{1}{c}{\multirow{2}{*}{Model}} & \multicolumn{2}{c}{Parameters} \\
\cline{2-3} \noalign{\smallskip}
& \multicolumn{1}{c}{Total} & \multicolumn{1}{c}{Trainable} \\
\midrule
VGG-16 & 16.55M & 17.77k \\
DenseNet-121 & 8.40M & 104.04k \\
ResNet-50 & 28.78M & 86.86k \\
\bottomrule
\end{tabular*}
\end{table}

\section{Discussion}

The experimental results show that Taylor-FO is an effective importance score. Without basis pruning, Taylor-FO outperformed L1 and HRank in 7 out of 9 combinations. This is probably because Taylor-FO utilizes information from both the feature maps and the weights, whereas L1 only depends on weight magnitudes and HRank only depends on feature maps. On the other hand, Taylar-FO and basis pruning outperformed each other in half of the experiments, while double pruning with Basis + Taylor-FO was dominant. This shows the appealing property of our double pruning approach that it can amplify existing pruning algorithms.

Although our main focus is not on domain adaptation, the experimental results show that our framework can decently transfer knowledge learned from ImageNet to other domains. ImageNet contains color images which are visually similar to the color images in CIFAR-10. In contrast, Fashion-MNIST contains grayscale images of fashion categories which are less similar, and MNIST contains grayscale handwritten digits which can be considered as a distant domain. Regardless of the domain similarity, our framework can transfer the knowledge from ImageNet with largely reduced model parameters and FLOPs. This is probably because the BN layers, which are important for domain adaptation \cite{Workshop:Li:ICLRWorkshop2017:Revisiting}, are trainable during transfer learning. In fact, we found that trainable BN layers are essential for good results.

As basis pruning does not change the original numbers of input and output channels of a convolutional layer, it can be applied to more complicated network architectures such as U-Net and V-Net for image segmentation \cite{Conference:Ronneberger:MICCAI2015,Conference:Milletari:3DV2016}. In contrast, double pruning introduces more complicated layer interactions thus further adjustments may be required. The performance and adjustments of the proposed framework on more complicated tasks can be studied in the future.

\section{Conclusion}

In this paper, we present the basis and double pruning approaches for efficient transfer learning. Using singular value decomposition, a convolutional layer can be decomposed into two consecutive layers with the basis vectors as their convolutional weights. With the basis scaling factors introduced, the basis vectors can be fine-tuned and pruned to reduce the network size and inference time, regardless of the existence of skip connections. Basis pruning can be further combined with other pruning algorithms for double pruning to obtain pruning ratios that cannot be achieved by either alone. Experimental results show that basis pruning outperformed pruning in the original feature space, and the performance was even more distinctive with double pruning.

\bibliographystyle{elsarticle-num}
\bibliography{Ref}

\end{document}